\documentclass[11pt,a4paper]{article}
\usepackage[margin=1in]{geometry} 
\usepackage{newtxtext,newtxmath}  

\usepackage[T1]{fontenc}
\usepackage[utf8]{inputenc}
\usepackage{microtype}

\usepackage{amsmath}
\DeclareMathOperator*{\E}{\mathbb{E}}

\usepackage{wrapfig}
\usepackage{needspace}
\usepackage{graphicx}
\usepackage{placeins}
\usepackage{booktabs}
\usepackage{multirow}
\usepackage{colortbl}
\usepackage{caption}
\usepackage{subcaption}
\usepackage{wrapfig}
\usepackage{pifont}
\usepackage{array}

\usepackage{xcolor}
\usepackage{changes}
\usepackage[accsupp]{axessibility} 
\usepackage[hidelinks]{hyperref}

\begin{document}

\begin{center}
{\LARGE\bfseries
MCPO: Mastery-Consolidated Policy Optimization for Large\\[0.2em]
Reasoning Models
\par}

\vspace{1.5em}

{\large
Zhaokang Liao$^{1,*}$ \quad
Yingguo Gao$^{1,*}$ \quad
Yi Yang$^{1,2}$ \quad
Yongheng Hu$^{1}$ \quad
Jingting Ding$^{1,\dagger}$
\par}

\vspace{0.55em}

{\large
$^{1}$Ant Group  \quad $^{2}$Zhejiang University
\par}

\vspace{0.9em}

{\small\ttfamily
lzk9508@mail.ustc.edu.cn \quad
yggaoeecs@gmail.com \quad
yang-yi@zju.edu.cn\\[0.25em]
yongheng.hyh@antgroup.com \quad
yimou.djt@antgroup.com
\par}

\vspace{0.4em}

\end{center}
\begingroup
\renewcommand{\thefootnote}{\fnsymbol{footnote}}
\footnotetext[1]{Equal contribution.}
\footnotetext[2]{Corresponding author: \texttt{yimou.djt@antgroup.com}.}
\endgroup
\setcounter{footnote}{0}
\vspace{1.0em}

\begin{abstract}
Reinforcement Learning with Verifiable Rewards (RLVR) has emerged as a promising approach to improve the reasoning abilities of Large Language Models (LLMs). Among RLVR algorithms, Group Relative Policy Optimization (GRPO) and its variants have demonstrated strong performance and high training efficiency. 
However, GRPO-style objectives exhibit two issues on high accuracy prompts including \textbf{\textit{mastered prompts}} (rollout accuracy \(=1\)) and \textbf{\textit{majority-correct  prompts}} (rollout accuracy \( \in (0.5,1) \)). 
For mastered prompts, group-relative advantages vanish, yielding no training signal and unconstrained policy drift that can cause forgetting. 
For majority-correct prompts, the induced query weight shrinks as accuracy increases, weakening consolidation from partial correctness to mastery. To alleviate this, we propose \textit{\textbf{M}astery-\textbf{C}onsolidated \textbf{P}olicy} \textit{\textbf{O}ptimization} (\textbf{MCPO}), which introduces (i) a hinge-KL regularizer applied exclusively to mastered prompts to bound harmful policy drift between successive gradient steps, and (ii) a weighting mechanism that prioritizes majority-correct prompts to  better allocate optimization effort. Extensive experiments across three mathematical benchmarks demonstrate that MCPO consistently improves pass@1 performance. Counter-intuitively, rather than restricting exploration, MCPO boosts pass@k metrics, indicating that mastery consolidation further catalyzes solution diversity. 

\end{abstract}

\section{Introduction}

The remarkable success of Large Reasoning Models (LRMs), such as OpenAI o1 \cite{jaech2024openai} and DeepSeek-R1 \cite{guo2025deepseek}, has sparked a paradigm shift in enhancing the reasoning capabilities of Large Language Models (LLMs). This shift is primarily driven by Reinforcement Learning with Verifiable Rewards (RLVR), which leverages automatically computed deterministic feedback (e.g., mathematical correctness or code-execution outcomes) for training. Compared with Reinforcement Learning from Human Feedback (RLHF) \cite{ouyang2022training,schulman2017proximal}, RLVR reduces dependence on costly human annotations and mitigates reward hacking\cite{amodei2016concrete,gao2023scaling}, a common failure mode of learning-based reward models, thereby providing a scalable and effective path toward autonomous improvement of the policy model. Among RLVR-based methods, Group Relative Policy Optimization (GRPO) \cite{deepseekmath} has emerged as a key advancement. By computing advantages using an intra-group average reward baseline instead of relying on a separate critic model\cite{stiennon2020learning} to estimate the value function\cite{mnih2015human}, GRPO improves training efficiency and performance in reasoning tasks, which has sparked growing interest in further improvements.
\label{sec:intro}
\begin{figure}[htbp]
  \centering
  \includegraphics[width=\linewidth]{figure1.pdf}

  \caption{Difference between DAPO and MCPO. We introduce a hinge KL loss term only applied on mastered prompts between policies of adjacent gradient steps and redesign the advantages for query weights reallocation.}
  \label{fig:fig1}

\end{figure}
\vspace{0.5em}
\newline
Numerous variants of GRPO have been proposed to strengthen exploration, training stability, and sample efficiency. For instance, DAPO \cite{dapo} introduces an asymmetric clipping scheme with
distinct thresholds, which helps prevent premature entropy collapse and thus promotes exploration. Meanwhile, GSPO \cite{zheng2025group}
reformulates importance sampling weights\cite{schulman2017proximal} and clipping at the sequence-level to improve stability and efficiency. DCPO \cite{yang2025dcpodynamicclippingpolicy}
and SAPO \cite{chen2026saposelfadaptiveprocessoptimization} further refine the ratio-clipping rule together with its update-scaling mechanism, alleviating zero or unstable
gradients induced by token-level clipping and reward standardization. Despite these advancements, a critical challenge remains largely overlooked: 
newly integrated skills often lack stability and remain susceptible to interference during iterative training.
\vspace{0.5em}
\newline 
Intuitively, the acquired knowledge and skills are primarily reflected in prompts whose rollout accuracy exceeds \(50\%\), including those with all rollouts correct (termed as \emph{mastered prompts}) and those that are majority-correct (\(0.5<p(x)<1\)). For mastered prompts, the advantages of all rollout responses collapse to zero, providing no gradient signal and allowing the policy to drift without constraint. 
As shown in Fig.~\ref{fig:mastered_prompts_curve}, mastered prompts quantitatively constitute up to one quarter of the prompts in a training batch. More importantly, mastered prompts are volatile: under the next step policy, the mean rollout accuracy of current step mastered prompts drops to 95\% on average across all training steps. This drop indicates a significant policy shift due to the absence of anchoring gradients\cite{ryu2019ode}. 
Beyond the policy drift on mastered prompts, the majority-correct prompts suffer from weight misallocation of GRPO-style objectives: the induced query weight peaks near \(p(x)=0.5\), while most training prompts are instead very easy or very hard. Even though majority-correct prompts occupy a large fraction in each batch, their query weights monotonically shrink as rollout accuracy increases, hindering the consolidation of partially learned skills into full mastery. 
These observations motivate us to explicitly optimize mastered prompts to prevent drift and reallocate weights for majority-correct prompts to consolidate partially learned skills.
\vspace{0.5em}
\newline
Based on the above analysis, we propose Mastery-Consolidated Policy Optimization (MCPO), which consolidates previously acquired knowledge from two complementary strategies: (i) Hinge-KL Regularization: for mastered prompts, we apply a hinge-KL constraint between the current policy and the old one on the last gradient step. This acts as a "knowledge anchor", bounding harmful policy drift and ensuring the model retains previously acquired reasoning skills. (ii) Query Weight Reallocation: for majority-correct prompts, MCPO adaptively upweights their contribution. Query weight reallocation mitigates the difficulty bias problem\cite{li2025disco} inherent in group relative advantages, ensuring majority-correct prompts provide a substantial learning signal before reaching full mastery.
\vspace{0.5em}
\newline
We evaluated MCPO on three standard mathematical reasoning benchmarks. Our results demonstrate consistent improvements in both the metrics $Pass@1$ and $Pass@k$\cite{chen2021evaluating}, confirming that consolidating acquired knowledge not only improves accuracy but also catalyzes further exploration.
\vspace{0.5em}
\newline
Our main contributions are summarized as follows:
\begin{itemize}
    \item We identify and analyze the phenomenon of performance degradation on mastered prompts, underscoring the need for mastery consolidation.
    \item We propose MCPO, a consolidation-based methods that effectively addresses the issue of no learning signal from mastered prompts and prevents the degradation of previously learned skills.
    \item We validate our method across three widely-recognized reasoning datasets. The results show that MCPO delivers substantial performance improvements, confirming its effectiveness for large reasoning models.
\end{itemize}

\section{Related Work}
\par
\textbf{Reinforcement Learning with Verifiable Rewards.}
Reinforcement Learning with Verifiable Rewards (RLVR) has become a key technique for improving the reasoning capabilities of large language models \cite{openai2024gpt4technicalreport,deepseekai2025deepseekv3technicalreport,yang2025qwen3technicalreport}. RLVR leverages deterministic reward signals, such as mathematical equivalence checking or program execution, to provide scalable and verifiable feedback without human preference labels or reward models.
Recent large reasoning models \cite{jaech2024openai,guo2025deepseek,kimiteam2026kimik25visualagentic,glm5team2026glm5vibecodingagentic}, such as DeepSeek-R1 \cite{guo2025deepseek} and Kimi-K2 \cite{team2025kimi}, highlight the effectiveness of large scale RLVR post-training.
One of the most representative approaches of RLVR is Group Relative Policy Optimization (GRPO) \cite{deepseekmath}. GRPO methods are widely used because they eliminate the critic model and the reward model, leading to markedly improved training efficiency and stability. A considerable body of literature has subsequently refined GRPO-based objectives to enhance exploration \cite{dapo,cui2025entropy,wu2025quantile,liang2025beyond,chen2025passktrainingadaptivelybalancing}, training stability \cite{zheng2025group,chen2026saposelfadaptiveprocessoptimization,zhao2026geometricmean}, and sample efficiency \cite{dapo,le2025no,nan2025ngrpo,li2025disco,liu2025understanding}, collectively shaping and enriching the broader paradigm of RLVR.
\vspace{0.5em}
\newline
\textbf{Enhancing Exploration via Entropy Control.}
A prevalent challenge in vanilla GRPO is the collapse of policy entropy\cite{haarnoja2018soft} during training, which hinders the exploration ability of the model. Wang et al. \cite{wang2025beyond} improve the RLVR by restricting gradient updates to high entropy tokens, presenting a pioneering investigation of RLVR from the perspective of token-level entropy dynamics. DAPO \cite{dapo} introduces the clip-higher technique, adopting a pair of asymmetric clipping parameters to mitigate the entropy collapse. Cui et al. \cite{cui2025entropy} reveals that the collapse of entropy is mechanically driven by the positive covariance between the probability of policy and their advantages, proposing two techniques termed Clip-Cov and KL-Cov to mitigate the issue of entropy collapse. QAE \cite{wu2025quantile} replaces
the mean reward with a group-wise K-quantile reward for the group baseline and shows that such a technique can help to avoid both entropy collapse and entropy explosion. Liang et al. \cite{liang2025beyond} propose an online self-play with variational problem synthesis (SvS), which requires the model to synthesize variational similar problems according to the correct solutions.
\vspace{0.5em}
\newline
\textbf{Sample Efficiency of GRPO Optimization.}
Zero variance samples and difficulty bias are two major sample efficiency problems of GRPO. Zero variance samples are defined as prompts where all generated responses within a sampled group receive identical rewards, all correct or all incorrect, leading to zero advantage and providing no training signal. Common practice in RLVR is to filter such prompts, such as dynamic sampling in DAPO \cite{dapo}, trading off compute efficiency against potential information loss. Recent concurrent work argues that zero-variance prompts can be exploited rather than discarded. RL-ZVP \cite{le2025no} proposes an entropy-guided token-level advantage shaping to activate optimization on all correct groups and all wrong ones, extracting learning signals even without within-group reward variance. NGRPO \cite{nan2025ngrpo} introduces a virtual positive sample to homogeneously incorrect groups to transfer the zero advantage to negative advantage and allow optimizing in all wrong groups. Difficulty bias, identified by Wang et al. in Dr-GRPO \cite{liu2025understanding}, means that prompts of different difficulty levels measured by rollout accuracy have unequal impacts on policy updates. Dr-GRPO removes the length and standard deviation
 normalization terms to achieve better token efficiency. DisCO \cite{li2025disco} theoretically analyzes the difficulty bias and provides a discriminative form of GRPO objective, decoupling the query weight from the discriminative term.


\section{Preliminaries}

This section begins by establishing the problem formulation and notation for RLVR, followed by a comprehensive review of two representative methods in this domain: GRPO and DAPO.
\vspace{0.5em}
\newline
\textbf{Problem Formulation and Notation.}
Let $x \sim \mathcal{D}$ denote a prompt sampled from a training distribution $\mathcal{D}$. A parameterized policy (i.e., a language model) $\pi_\theta$, parameterized by $\theta$, generates a response sequence $y = (y_1, \dots, y_T)$ with the joint probability defined as $\pi_\theta(y \mid x) = \prod_{t=1}^{T} \pi_\theta(y_t \mid x, y_{<t})$. To evaluate the quality of this response, a verifiable reward function $R(x, y)$ maps a prompt-response pair to a scalar reward $R \in \mathbb{R}$. For each prompt $x$, we sample a rollout group consisting of $G$ independent responses from an old policy model $\pi_{\theta_{\mathrm{old}}}$, denoted as $\{y_i\}_{i=1}^{G} \sim \pi_{\theta_{\mathrm{old}}}(\cdot \mid x)$, and compute the corresponding rewards $\{R_i\}_{i=1}^{G}$. These rewards are then used to derive advantage estimates, which serve as learning signals to optimize the policy model $\pi_\theta$.
\vspace{0.5em}
\newline
\textbf{Group Relative Policy Optimization (GRPO).}
GRPO eliminates the need for a learned value function by estimating advantages directly from the reward distribution within each rollout group. Concretely, given a prompt $x$ and a rollout group $\{y_i\}_{i=1}^{G}$ with corresponding rewards $\{R_i\}_{i=1}^{G}$, the advantage for the $i$-th response is defined as:
{
\begin{equation}
    A_i = \frac{R_i - \mathrm{mean}(\{R_j\}_{j=1}^{G})}{\mathrm{std}(\{R_j\}_{j=1}^{G})},
\end{equation}
}
\vspace{0.5em}
\newline
where $\mathrm{mean}(\cdot)$ and $\mathrm{std}(\cdot)$ denote the empirical mean and standard deviation over the group. 
The policy optimization follows a PPO-style objective\footnote{For clarity of exposition, we omit the KL-penalty term.} with clipped importance sampling:
{
\begin{equation}
\mathcal{J}_{\mathrm{GRPO}}(\theta)
=
\E_{x \sim \mathcal{D}}
\Bigg[
\frac{1}{G}\sum_{i=1}^{G}\frac{1}{|y_i|}\sum_{t=1}^{|y_i|}
\min\Bigl(
r_{i,t} A_i,\ 
\mathrm{clip}\!\bigl(r_{i,t},\,1-\epsilon,\,1+\epsilon\bigr)\,A_i
\Bigr)
\Bigg].
\end{equation}
}
\vspace{0.5em}
\newline
where $r_{i,t} = \pi_\theta(y_{i,t} \mid x, y_{i,<t}) / \pi_{\theta_{\mathrm{old}}}(y_{i,t} \mid x, y_{i,<t})$ denotes the importance sampling ratio at the token position $t$, and $\epsilon$ controls the trust-region constraint imposed on the policy update.
\vspace{0.5em}
\newline
\textbf{Dynamic Sampling Policy Optimization (DAPO).}
DAPO can be regarded as an improved variant of GRPO that incorporates several deliberate refinements. In particular, DAPO omits the KL-penalty term and adopts an asymmetric clipping interval $(1-\epsilon_{\mathrm{low}},\,1+\epsilon_{\mathrm{high}})$, allowing more aggressive updates for positive advantage actions. Moreover, the objective is normalized at the token-level as follows:
{
\begin{equation}
\mathcal{J}_{\mathrm{DAPO}}(\theta)
=
\E_{x \sim \mathcal{D}}
\Bigg[
\frac{1}{\sum_{i=1}^{G} |y_i|}
\sum_{i=1}^{G}\sum_{t=1}^{|y_i|}
\min\Bigl(
r_{i,t} A_i,\ 
\mathrm{clip}\!\bigl(r_{i,t},\,1-\epsilon_{\mathrm{low}},\,1+\epsilon_{\mathrm{high}}\bigr)\,A_i
\Bigr)
\Bigg].
\end{equation}
}
\vspace{0.5em}
\newline
DAPO further introduces a dynamic sampling constraint:
{
\[
1 \le \sum_{i=1}^{G} \mathbf{1}\!\left\{ R(x,y_i)=1 \right\} \le G-1 .
\]
}
\newline
This constraint filters out prompts whose rollout groups are either uniformly correct or uniformly incorrect. We choose DAPO as the baseline to compare with our proposed method due to its competitive performance and broad adoption in recent RLVR studies.

\section{Observations}
\label{issue}
In this section, we present our empirical observations on DAPO training dynamics, focusing on two aspects: policy drift on mastered prompts and weight misallocation across prompts with different difficulties.
\subsection{Policy Drift on Mastered Prompts}
\label{sec:observation}
\begin{figure}[t]
  \centering
  \begin{subfigure}[t]{0.49\linewidth}
    \centering
    \includegraphics[width=\linewidth]{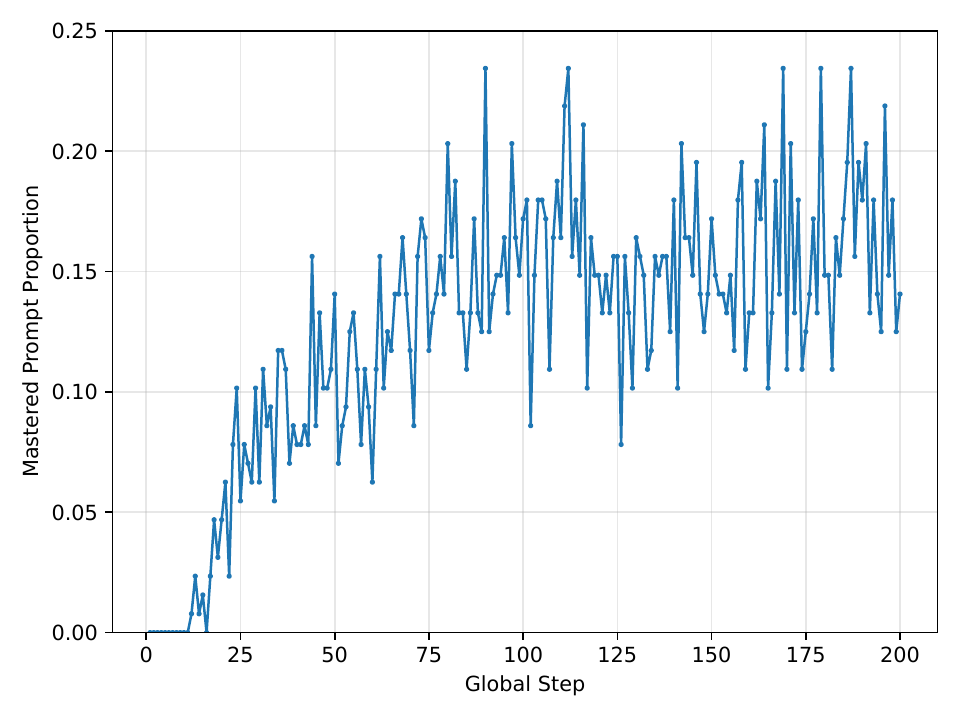}

    \caption{Mastered prompt proportion at different global steps.}
    \label{fig:mastered_prompts_curve}
  \end{subfigure}\hfill
  \begin{subfigure}[t]{0.49\linewidth}
    \centering
    \includegraphics[width=\linewidth]{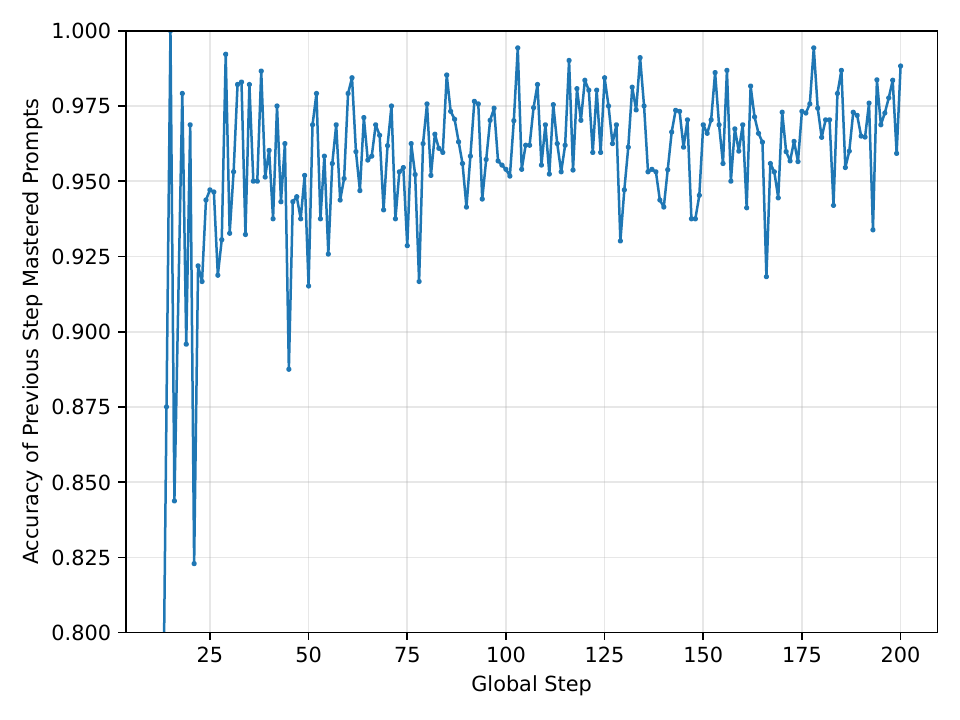}
    \caption{Mean rollout accuracy of previous step mastered prompts under current policy.}
    \label{fig:prev_mastered_prompts_accuracy_curve}
  \end{subfigure}
  \caption{Dynamics of mastered prompts in different global step during training. Each global step contains a training batch of 128 prompts.}
  
  \label{fig:mastery_dynamics}
\end{figure}

This subsection presents empirical observations regarding policy drift on mastered prompts. In the GRPO method and its variants, mastered prompts suffer from a vanishing learning signal. Since the within-group relative advantages are calculated based on reward variance, an all correct rollout group causes these advantages to collapse to zero. Consequently, the advantage-weighted gradient becomes zero, rendering these prompts effectively inactive in the optimization objective.
\vspace{0.5em}
\newline
While prior works \cite{dapo,zheng2025actpaysefficientreinforcement} employ online filtering to discard such signal-free prompts to improve training efficiency, we hypothesize that continued optimization on the remaining prompts may inadvertently lead to decline of previously mastered skills. To study this policy drift, we conduct the following experiment: we disable the online filter and evaluate the mean rollout accuracy on mastered prompts under the policy of the immediate preceding global step. Crucially, these rollouts serve purely as a diagnostic probe and do not contribute to gradient updates. Our experiments are conducted using the Qwen3-8B-Base model \cite{yang2025qwen3technicalreport} on the DAPO-17K dataset \cite{dapo}, with a training batch size of 128 and 16 rollouts per prompt. At each global step, we record (i) the number of mastered prompts identified at that step, and (ii) the mean rollout accuracy on the set of mastered prompts in the previous global step under the current policy.
\vspace{0.5em}
\newline
Experimental results confirm that models are susceptible to losing skills acquired earlier in the training process. Fig.~\ref{fig:mastered_prompts_curve} illustrates the trajectory of the proportion of mastered prompts across global steps. As shown in Fig.~\ref{fig:mastered_prompts_curve}, the fraction of mastered prompts continues to increase during training, ultimately accounting for approximately 25\%.  Fig.~\ref{fig:prev_mastered_prompts_accuracy_curve} depicts the rollout accuracy of the current policy on mastered prompts in the preceding step. Notably, the accuracy on these previously mastered queries consistently regresses to approximately 95\% within a single global update, with pronounced fluctuations appearing in the early stages of training. This rapid one-step regression is both undesirable and counterintuitive. In human cognition, practicing new tasks rarely causes an immediate erosion of recently solidified skills.
\vspace{0.5em}
\newline
We attribute this behavior to a lack of mastery consolidation in GRPO-like algorithms. Specifically, the zero-advantage nature of mastered prompts provides no explicit gradient-based mechanism to preserve existing behaviors. Simultaneously, updates driven by non-mastered prompts—which may involve high-variance gradients or localized overfitting—can cause the policy to drift away from the optimal manifold for mastered queries.


\subsection{Issue of Weight Misallocation}
\label{difficultybias}
\begin{figure}[t]
  \centering
  \begin{subfigure}[t]{0.56\linewidth}
    \centering
    \includegraphics[width=\linewidth]{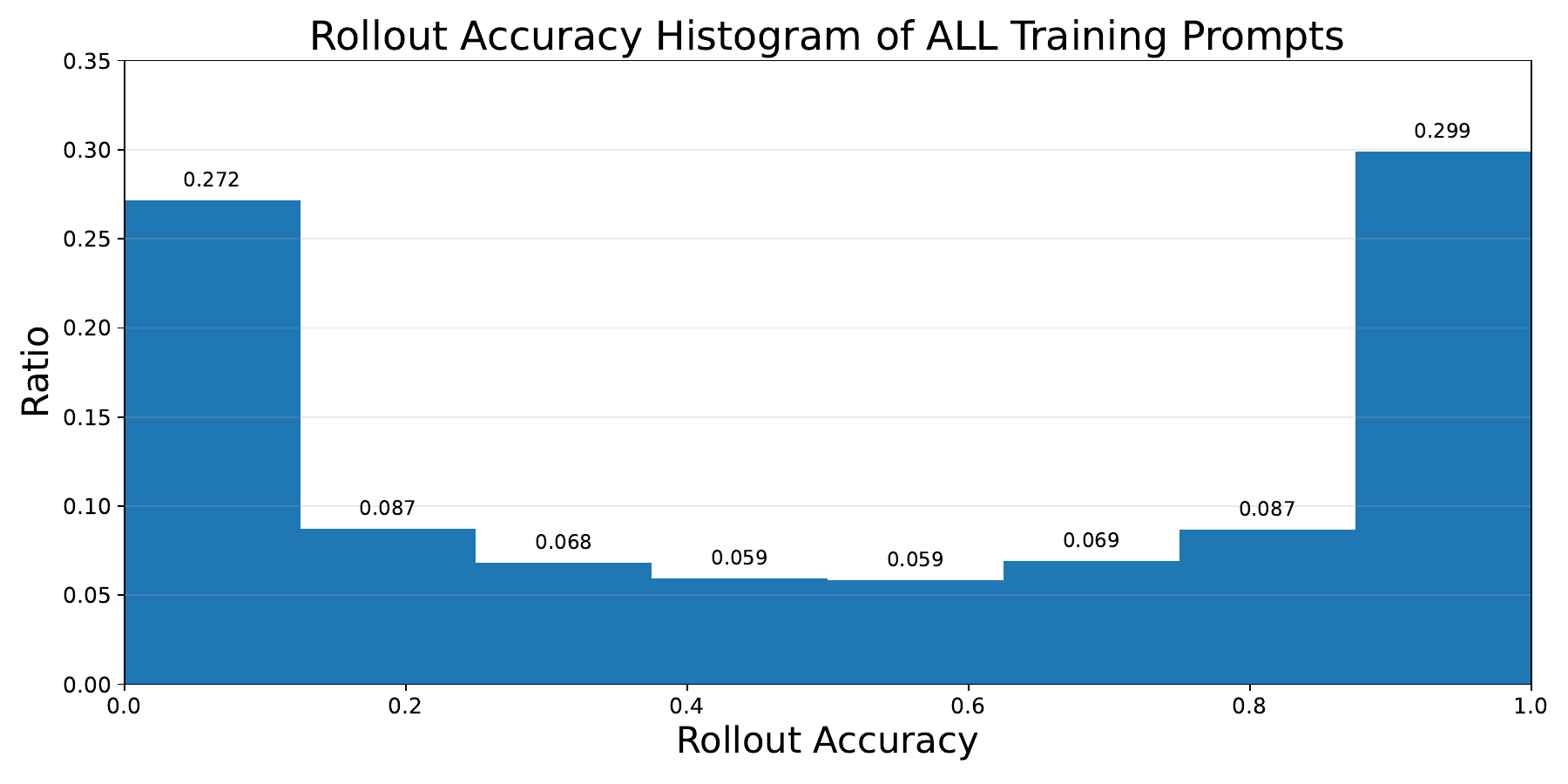}
    \caption{The statistics of rollout accuracy histogram among all training prompts.}
    \label{fig:hist}
  \end{subfigure}\hfill
  \begin{subfigure}[t]{0.43\linewidth}
    \centering
    \includegraphics[width=\linewidth]{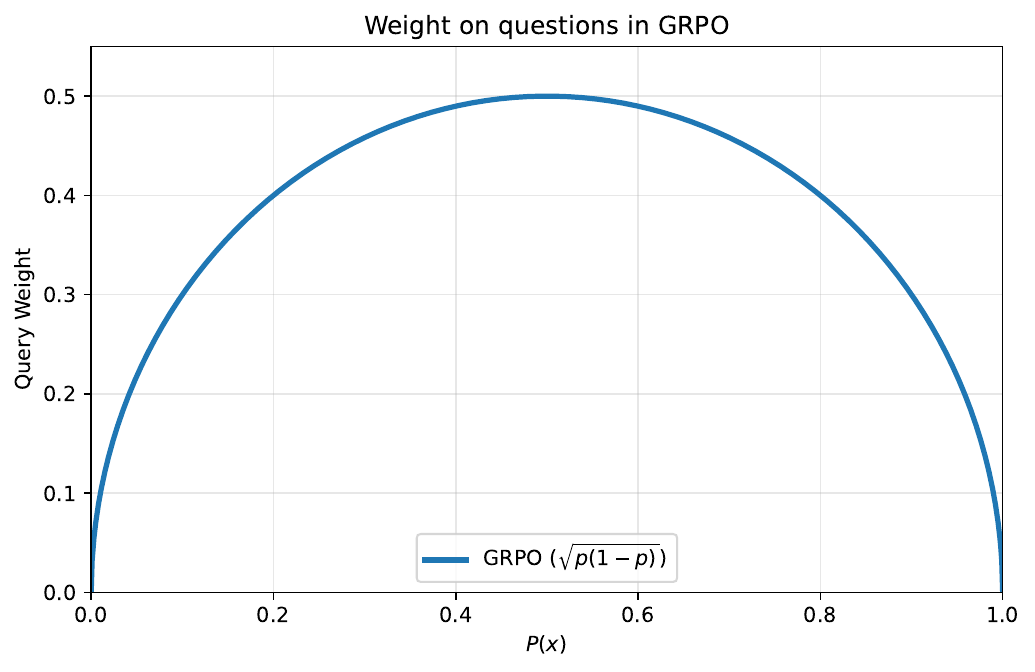}
    \caption{GRPO query weight as a function of rollout accuracy \(p(x)\).}
    \label{fig:fig2}
  \end{subfigure}
  \caption{Rationale and mechanism of weight allocation in GRPO. (a) Histogram of rollout accuracy over all training prompts, showing a bimodal distribution with a substantial fraction of prompts at very low and very high accuracies. (b) GRPO query weight as a function of rollout accuracy $p(x)$, which peaks at $p(x)=0.5$ and decreases to zero as $p(x)$ approaches 0 or 1, thus emphasizing prompts of intermediate difficulty.}
  \label{fig:query_weight_compare}
  
\end{figure}

Beyond the drift on mastered prompts, we observe another inefficiency in GRPO-style objectives:
the implicit query weight induced by group-relative advantages can be severely mismatched
with the empirical distribution of prompt difficulties.
This phenomenon is closely related to the difficulty bias~\cite{li2025disco,liu2025understanding},
which refers to a systematic imbalance in how prompts of different rollout precisions contribute to the overall policy update.
DisCO\cite{li2025disco} offers a discriminative perspective of GRPO objective to theoretically analyze the difficulty bias, by decoupling a
query-level weight from a discriminative term. Consider binary verifiable rewards where the reward for the correct response is 1 and the incorrect one is 0, the 
precision $p(x)$ in the response group for  prompt $x$ is defined as
{
\begin{equation}
p(x)= \frac{1}{G}\sum_{i=1}^{G}\mathbf{1}\!\left\{ R(x,y_i)=1 \right\},
\qquad y_i \overset{\text{i.i.d.}}{\sim} \pi_{\theta_{\mathrm{old}}}(\cdot\mid x).
\label{eq:rollout_accuracy}
\end{equation}
}
\vspace{0.5em}
\newline
Let $\pi^{+}_{\theta_{\mathrm{old}}}(\cdot\mid x)$ and $\pi^{-}_{\theta_{\mathrm{old}}}(\cdot\mid x)$ denote
the old-policy distributions on the correct and incorrect response, respectively. Token-wise importance sampling ratio is aggregated along each response sequence with length normalization and is denoted as $s^{+}_{\theta}(y^{+},x)$ for correct response or $s^{-}_{\theta}(y^{-},x)$ for incorrect response. In GRPO, the distinction between $s^{+}$ and $s^{-}$ arises from the clipping-based surrogate as follow definition: 
{
\begin{equation}
\begin{aligned}
s^{+}_{\theta}(y^{+},x)
&= \frac{1}{|y^{+}|}\sum_{t=1}^{|y^{+}|}
f^{+}\!\left(
\frac{\pi_{\theta}\!\left(y^{+}_t \mid x, y^{+}_{<t}\right)}
     {\pi_{\theta_{\mathrm{old}}}\!\left(y^{+}_t \mid x, y^{+}_{<t}\right)},
\,1\right),\\
s^{-}_{\theta}(y^{-},x)
&= \frac{1}{|y^{-}|}\sum_{t=1}^{|y^{-}|}
f^{-}\!\left(
\frac{\pi_{\theta}\!\left(y^{-}_t \mid x, y^{-}_{<t}\right)}
     {\pi_{\theta_{\mathrm{old}}}\!\left(y^{-}_t \mid x, y^{-}_{<t}\right)},
\,1\right).
\end{aligned}
\end{equation}
}
\vspace{0.5em}
\newline
where $f^{+}(r,1)=\min(r,1+\epsilon)$, $f^{-}(r,1)=\max(r,1-\epsilon)$.
%
Then the GRPO objective can be derived in the following form~\cite{li2025disco}:
{\footnotesize
\begin{equation}
\mathcal{J}_{\mathrm{GRPO}}(\theta)
=
\E_{x\sim\mathcal{D}}
\left[
\underbrace{\sqrt{p(x)\bigl(1-p(x)\bigr)}}_{\text{query weight}}
\cdot
\underbrace{
\mathbb{E}_{y^{+}\sim \pi^{+}_{\theta_{\mathrm{old}}},\, y^{-}\sim \pi^{-}_{\theta_{\mathrm{old}}}}
\!\left[
s^{+}_{\theta}(y^{+},x)-s^{-}_{\theta}(y^{-},x)
\right]
}_{\text{discriminative term}}
\right].
\label{eq:disco_query_weight}
\end{equation}
}
\vspace{0.5em}
\newline
From Eq.~\eqref{eq:disco_query_weight}, the GRPO objective contains a query-level weight that depends only on the prompt difficulty measured by the precision of the rollout $p(x)$. As shown in Fig.~\ref{fig:fig2}, prompts of intermediate difficulty receive the largest query weight, while very easy and very hard prompts are downweighted.
\vspace{0.5em}
\newline
Furthermore, we record the statistics of the rollout accuracy histogram of all training prompts shown in Fig.~\ref{fig:hist}. Throughout the training process, we observe that prompts with a rollout accuracy around 
0.5 constitute only a small fraction of the total training prompts. In contrast, most of the training prompts exhibit either very low or very high rollout accuracy. Such a rollout accuracy distribution is the inverse of the query weight distribution. As a result, the policy assigns disproportionately large update weights to the small subset of moderately difficult prompts, while downweighting the majority of prompts. This constitutes a severe misallocation of the query weight. 
\vspace{0.5em}
\newline
Consequently, although majority-correct prompts (with rollout accuracy above 
50\%) account for a substantial portion of the training data, GRPO assigns them progressively smaller query weights as their accuracy increases, thereby attenuating their influence on the policy update and making it harder for the model to consolidate partially learned knowledge. 

\section{Mastery Consolidated Policy Optimization}
In this section, we introduce our method, Mastery Consolidated Policy Optimization (MCPO), to address the issues mentioned in Section~\ref{issue}.
We present the two core components of MCPO: (i) a hinge-KL penalty term that anchors the model's behavior on mastered prompts between adjacent step policies to address policy drift, and (ii) a query weight balancing strategy designed for majority-correct prompts to mitigate the difficulty bias. Together, these two mechanisms consolidate the acquired knowledge of the model.


\subsection{Consolidate with Hinge-KL Loss on Mastered Prompts}\label{subsec:hkl}
We observe that policy updates can induce substantial changes in the output distribution on mastered prompts, although mastered prompts contribute no effective learning signal.
This phenomenon is undesirable because excessive drift on mastered prompts erodes previously learned behaviors, leading to forgetting and performance regression.
Moreover, for mathematical and other reasoning tasks, the underlying knowledge is largely systematic and internally consistent. As a result, newly acquired knowledge should be compatible with the knowledge that has already been mastered.
From this perspective, learning from non-mastered prompts should not induce substantial shifts in the output distribution on mastered prompts.  Motivated by these considerations, we introduce an explicit consolidation mechanism that limits step-to-step policy drift on mastered prompts, thereby preserving already mastered performance. 
\vspace{0.5em}
\newline
Specifically, we augment the objective with a hinge-KL divergence term that constrains policy changes between adjacent update steps.
For each gradient update, we denote the pre-update policy as $\pi_{\mathrm{old}}$ and the post-update policy as $\pi_{\theta}$, and penalize their divergence using a hinge form with margin $\delta$.
In MCPO, we apply the hinge-KL term only to the mastered prompts subset, so that it functions as a consolidation constraint rather than competing with learning on non-mastered prompts.
Given a mastered prompt $x_b$ and a sampled response $y$, we define log-ratio form policy drift at token t as:
\begin{equation}
d_{t}(x,y)=
\log \pi_{\theta}\!\left(y_{t}\mid x,y_{<t}\right)
-
\log \pi_{\mathrm{old}}\!\left(y_{t}\mid x,y_{<t}\right),
\end{equation}
which measures the policy change between two consecutive gradient steps.
We used the $k_3$ estimator to approximate the reverse KL divergence. Therefore, the token-level reverse KL divergence between the current policy and the old policy can be approximated as: 
\begin{equation}
k_3(d_{t}(x,y))= e^{d_{t}(x,y)}-d_{t}(x,y)-1.
\end{equation}
We then impose a drift budget $\delta$  and define
\begin{equation}
c_{+}= k_3(\delta)=e^{\delta}-\delta-1,
\qquad
c_{-}= k_3(-\delta)=e^{-\delta}+\delta-1.
\end{equation}
The resulting hinge-KL penalty is
\begin{equation}
\phi(d_{t}(x,y))=
\begin{cases}
0, & |d_{t}(x,y)|\le \delta,\\
k_3(d_{t}(x,y))-c_{+}, & d_{t}(x,y)>\delta,\\
k_3(d_{t}(x,y))-c_{-}, & d_{t}(x,y)<-\delta.
\end{cases}
\end{equation}
We apply the hinge-KL penalty only on mastered prompts between current policy and the old one on the last gradient step. As a result, the resulting hinge-KL term is defined as:

\begin{equation}
\mathbb{D}_{\mathrm{HKL}}\!\Big(\pi_{\mathrm{old}}\,\Big\|\,\pi_{\theta}\Big)
=
\mathbb{E}_{x\in\mathcal{M}}
\
\mathbb{E}_{y\sim \pi_{\mathrm{old}}(\cdot\mid x)}
\left[
\frac{1}{|y|}\sum_{t=1}^{|y|}\phi\!\left(d_{t}(x,y)\right)
\right],
\label{eq:Dhkl_expand}
\end{equation}
\vspace{0.5em}
\newline
where $\mathcal{M}$ denote the set of mastered prompts. This hinge form tolerates small, benign step-to-step changes while penalizing large drift beyond the budget.

\subsection{Query Weight Balance on Majority-Correct Prompts}

\begin{wrapfigure}{r}{0.52\linewidth}
\centering

\includegraphics[width=\linewidth]{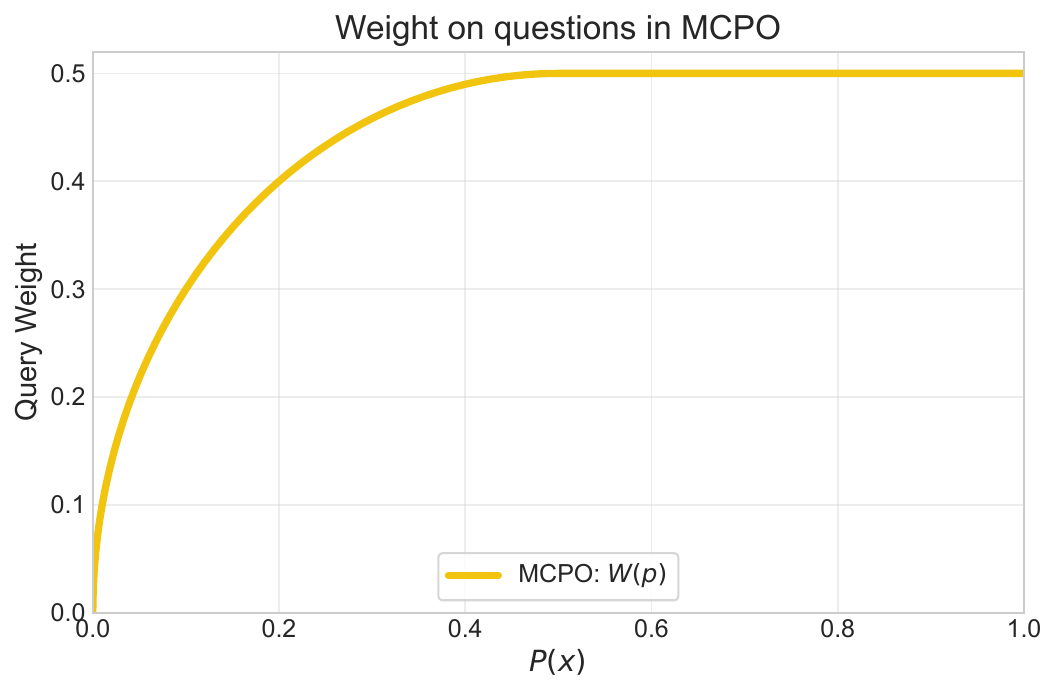}

\caption{Distribution of query weight associated with MCPO objective.}
\label{fig:mcpo_query_weight}

\end{wrapfigure}
\par As described in Section~\ref{difficultybias}, the training objective induces a difficulty bias that overweights the small minority of medium-accuracy prompts while underweighting the vast majority of very easy or very hard prompts, thereby bottlenecking optimization. To mitigate this issue, we adjust the group relative advantage estimation function so that the query weight remains constant where rollout accuracy exceeds 50\%. Concretely, given a prompt $x$ and a rollout group $\{y_i\}_{i=1}^{G}$ with corresponding rewards $\{R_i\}_{i=1}^{G}$, the advantage for the $i$-th response is defined as:
{
\begin{equation}
A_i^{\mathrm{MCPO}}(x)
=
\frac{R_i - \mathrm{mean}(\{R_j\}_{j=1}^{G})}
{\mathrm{std}(\{R_j\}_{j=1}^{G}) \cdot \mathrm{scale}\!\big(p(x)\big)} \,,
\label{eq:amcpo}
\end{equation}
}
\vspace{0.5em}
\newline
where \(p(x)\) is the rollout precision defined in Eq.~\eqref{eq:rollout_accuracy}\ and
{
\begin{equation}
\mathrm{scale}\!\big(p(x)\big)=
\begin{cases}
1, & p(x)\le 0.5,\\[4pt]
2\sqrt{p(x)\big(1-p(x)\big)}, & p(x) > 0.5.
\end{cases}
\end{equation}
}
\vspace{0.5em}
\newline
According to the group relative advantage defined in Eq.~\eqref{eq:amcpo}, we can derive the query weight of MCPO as follows:
{
\begin{equation}
W(x)=
\begin{cases}
\sqrt{p(x)\bigl(1-p(x)\bigr)}, & p(x)\le 0.5,\\[6pt]
0.5, & p(x)>0.5.
\end{cases}
\label{w:mcpo}
\end{equation}
}
\vspace{0.5em}
\newline
The relationship between \(W(x)\) and \(p(x)\) described in Eq.~\eqref{w:mcpo} is illustrated in Fig.~\ref{fig:mcpo_query_weight}.
Combining the hinge-KL term introduced in Section~\ref{subsec:hkl}, which constrains the step-to-step policy drift on mastered prompts, with the MCPO advantage defined in this section for query weight balance, the overall objective of discriminative form can be obtained as:

{
\begin{equation}
\begin{split}
\mathcal{J}_{\mathrm{MCPO}}(\theta)
=
\E_{x\sim\mathcal{D}}
\Big[
W(x)\cdot
\mathbb{E}_{y^{+}\sim \pi^{+}_{\theta_{\mathrm{old}}},\, y^{-}\sim \pi^{-}_{\theta_{\mathrm{old}}}}
\!\left[
s^{+}_{\theta}(y^{+},x)-s^{-}_{\theta}(y^{-},x)
\right]\Big]
-
\beta\;
\mathbb{D}_{\mathrm{HKL}}\!\Big(\pi_{\mathrm{old}}\,\Big\|\,\pi_{\theta}\Big).
\end{split}
\label{eq:total_obj_with_Dhkl}
\end{equation}
}

\section{Experiments}
In this section, we conduct a comprehensive empirical evaluation of the proposed MCPO. Although MCPO’s hinge-KL consolidation term and group relative advantage re-scaling are broadly compatible with GRPO and its variants, we adopt DAPO as our baseline for comparison, as it is one of the most widely used GRPO variants in academic literature and the query weight distribution between DAPO and GRPO is the same.

\subsection{Experiment Settings}
\textbf{Experiment Datasets and Evaluation Metrics.} We conduct experiments on mathematical reasoning tasks to evaluate our method. In particular, we use DAPO-Math-17K dataset for training, which consists of nearly 17k unique question-answer pairs. During validation, we evaluate the model on three standard mathematical reasoning benchmark: AIME 2024, AIME 2025, and AMC 2023. We adopt pass@1 metric averaged over $k=8$ samples using a rollout temperature of 0.7 and a max response length of 20480 to assess the model accuracy and pass@8 metric to quantify the exploration ability. All evaluations are conducted in zero-shot mode\cite{brown2020language}.
\vspace{0.5em}
\newline
\textbf{Models and Baseline Methods.} We conduct experiments on two base models: Qwen3-8B-Base and Qwen3-14B-Base \cite{yang2025qwen3technicalreport}. We adopt DAPO, one of the most popular RLVR algorithms, as baseline for comparison. Prior to RLVR, we did not perform any cold-start or other supervised fine-tuning. 
\vspace{0.5em}
\newline
\textbf{Training Details.} Our implementation is primarily based on the VeRL framework \cite{Sheng_2025}, with additional modifications. Except for the dynamic sampling with online filter, MCPO is trained under the same settings as DAPO. Both methods adopt techniques including clip higher, seq-mean-token-mean gradient loss aggregation mode and overlong shaping. We also apply identical hyperparameters for fair comparison. We utilize the AdamW\cite{kingma2014adam} optimizer with a constant learning rate of \(1.0 \times 10^{-6}\) and no learning rate warmup
or scheduling is adopted. We employ a training batch
size of 128 and a mini-batch size of 64. The model generates 16 responses for each question. For clipping setting, we set clipping parameter  $\epsilon_{\mathrm{low}}=0.2$ and $\epsilon_{\mathrm{high}}=0.28$. For overlong reward shaping, the maximum response length is 20480 and the punish cache length is 4096. For the unique hyperparameters of MCPO, we experimentally set the coefficient of hinge KL loss term $\beta=1$ and the policy drift tolerant budget on mastered prompts $\delta=0.01$.

\subsection{Main Comparison Results}
We evaluate two Qwen3 series base models across three popular mathematical benchmark to
demonstrate the effectiveness of MCPO. As shown in Table~\ref{tab:xxx}, MCPO consistently improves over DAPO on all three benchmarks, boosting both \textit{pass@1} and \textit{pass@8} performance metric. For Qwen3-8B-base model, MCPO achieves \textit{pass@1} performance gains of 4.58 points on AIME24 and 2.5 points on AIME25. MCPO still yields measurable \textit{pass@1} gains of 2.18 points on AMC23 where the baseline performance is already high. Beyond our expectations, MCPO also delivers \textit{pass@8} performance improvement of 5.29 points on AIME24 and 2.49 points on AIME25. On AMC23, MCPO achieves \textit{pass@8} performance comparable to that of DAPO, since AMC23 is so simple that the \textit{pass@8} performance of the baseline methods has already saturated. The fact that MCPO improves \textit{pass@8} simultaneously with \textit{pass@1} indicates that the method does not trade exploration for exploitation. Instead, MCPO increases the probability of discovering correct solutions within multiple samples, consistent with enhanced solution diversity.
\begin{table}[t]
\centering
\small
\vspace{-35pt}
\caption{Performance comparison between MCPO and DAPO on the Qwen3-8B-base model and Qwen3-14B -base model.}

\setlength{\tabcolsep}{10pt}
\renewcommand{\arraystretch}{1.25}
\resizebox{0.9\textwidth}{!}{
\begin{tabular}{l cc cc cc}
\toprule
\multirow{2}{*}{Method} &
\multicolumn{2}{c}{AIME24} &
\multicolumn{2}{c}{AIME25} &
\multicolumn{2}{c}{AMC23} \\
\cmidrule(lr){2-3}\cmidrule(lr){4-5}\cmidrule(lr){6-7}
& pass@1 & pass@8 & pass@1 & pass@8 & pass@1 & pass@8 \\
\midrule

\rowcolor{blue!15}
\multicolumn{7}{c}{\textbf{Qwen3-8B-Base Model}}\\
\midrule

DAPO \cite{dapo}
& 0.4000 & 0.6483
& 0.3083 & 0.4824
& 0.8438 & 0.9706 \\
MCPO (Ours)
& \textbf{0.4458} & \textbf{0.7012}
& \textbf{0.3333} & \textbf{0.5073}
& \textbf{0.8656} & \textbf{0.9837} \\
\midrule

\rowcolor{blue!15}
\multicolumn{7}{c}{\textbf{Qwen3-14B-Base Model}}\\
\midrule
DAPO \cite{dapo}
& 0.5292 & 0.7620
& 0.4041 & 0.5980
& 0.8937 & \textbf{0.9880} \\
MCPO (Ours)
& \textbf{0.5583} & \textbf{0.7753}
& \textbf{0.4292} & \textbf{0.6440}
& \textbf{0.9094} & 0.9806 \\
\bottomrule
\end{tabular}
}
\vspace{-5pt}
\label{tab:xxx}

\end{table}

\vspace{-20pt}
\subsection{Ablation Study}


\begin{table}[t]
\centering
\small

\renewcommand{\arraystretch}{1.12}
\caption{Ablation study of MCPO components on Qwen3-8B-Base. "Hinge-KL Loss" denotes the consolidation constraint on mastered prompts, and "Reweight" denotes the query-weight reallocation.}

\label{tab:ablation}
\resizebox{0.9\textwidth}{!}{
\begin{tabular}{@{}lcccccc@{}}
\toprule
\multirow{2}{*}{Method} &
\multicolumn{2}{c}{AIME24} &
\multicolumn{2}{c}{AIME25} &
\multicolumn{2}{c}{AMC23} \\
\cmidrule(lr){2-3}\cmidrule(lr){4-5}\cmidrule(lr){6-7}
& pass@1 & pass@8 & pass@1 & pass@8 & pass@1 & pass@8 \\
\midrule

\rowcolor{gray!15}\textbf{MCPO (Ours)} &
\textbf{0.4458} & \textbf{0.7012} &
0.3333 & \textbf{0.5073} &
\textbf{0.8656} & \textbf{0.9837} \\

\quad -- w/o Hinge-KL Loss &
0.4042 & 0.6563 & 0.3167 & 0.4881 &
0.8406 & 0.9716 \\

\quad -- w/o ReWeight &
0.4417 & 0.6812 &
\textbf{0.3375} & 0.5071 &
0.8625 & 0.9751 \\

\midrule
DAPO \cite{dapo} &
0.4000 & 0.6483 &
0.3083 & 0.4824 &
0.8438 & 0.9706 \\

\bottomrule
\end{tabular}
}

\end{table}
\begin{figure}[b!]
  \centering
  \begin{subfigure}[t]{0.49\linewidth}
    \centering
    \vspace{-10pt}
    \includegraphics[width=\linewidth]{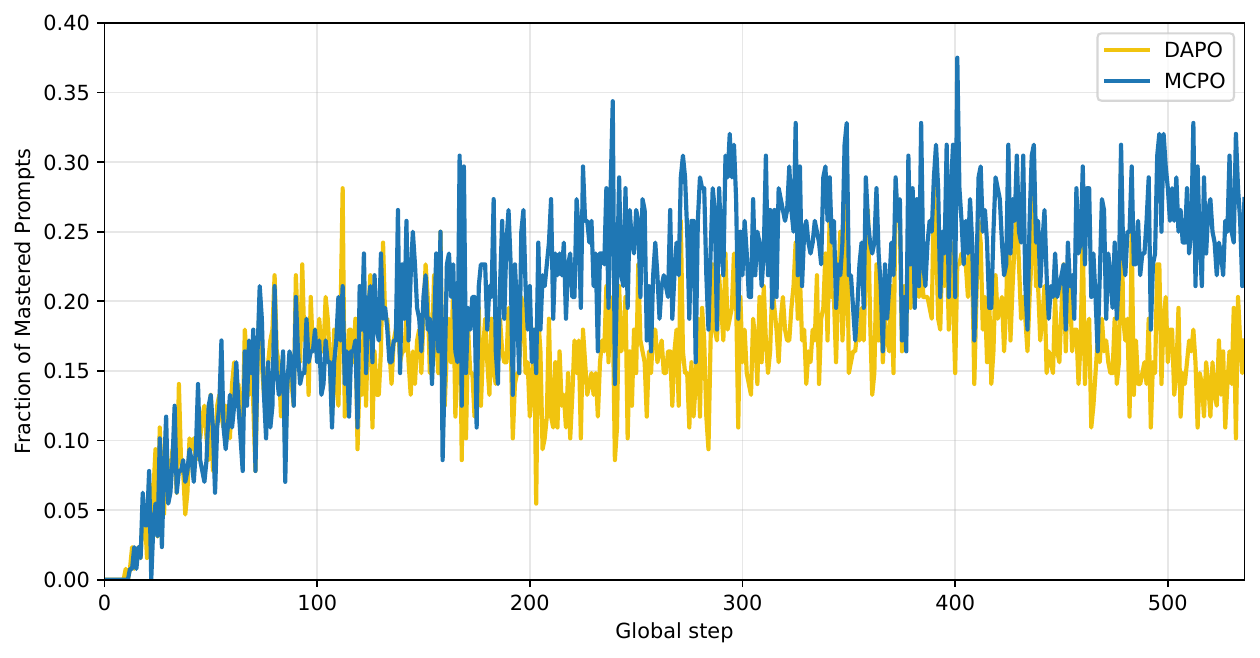}
    \caption{Fraction of mastered prompts in DAPO and MCPO.}
    \label{fig:mastered-frac}
  \end{subfigure}\hfill
  \begin{subfigure}[t]{0.49\linewidth}
    \centering
    \vspace{-10pt}
    \includegraphics[width=\linewidth]{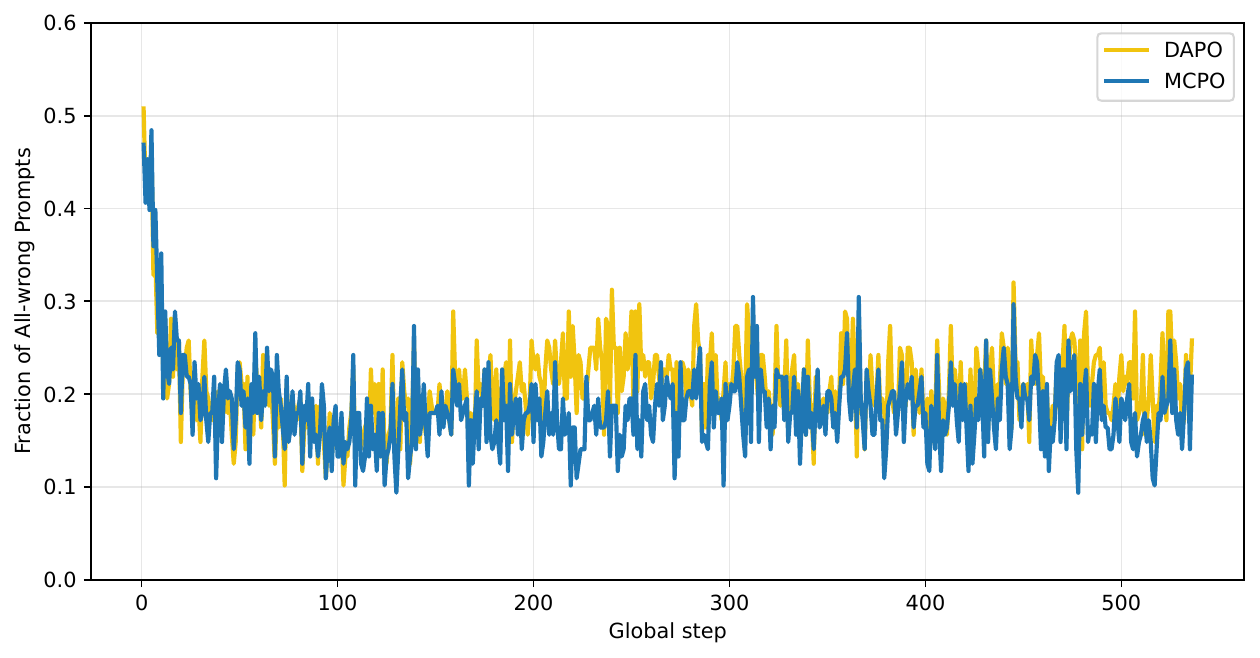}
    \caption{Fraction of all wrong prompts in DAPO and MCPO.}
    \label{fig:all-wrong-frac}
  \end{subfigure}
  \caption{Fractions of mastered and all wrong prompts in DAPO and MCPO. MCPO yields a higher fraction of mastered prompts and a lower fraction of all wrong prompts.}
  \label{fig:prompt-fractions}
  
\end{figure}
To disentangle the effects of the two components in MCPO, we conduct an ablation study on the Qwen3-8B-base model. The ablation results are reported in Table~\ref{tab:ablation}.
The baseline method, which lacks both of our proposed components, establishes the weakest performance across all benchmarks. Introducing hinge-KL loss term alone yields consistent gains on both \textit{pass@1} and \textit{pass@8} across all three benchmarks, indicating that constraining policy drift on mastered prompts is crucial for improving overall accuracy. In contrast, applying query weight balance alone provides comparable performance of \textit{pass@1} metrics and mainly benefits \textit{pass@8} performance, suggesting its primary effect is to increase multi-sample success. Combining both components delivers the best overall performance: MCPO achieves the strongest results on 5 out of 6 reported metrics, supporting the complementarity between hinge-KL loss term and query weight reallocation. On AIME25, MCPO is slightly lower than the hinge-KL loss only variant on \textit{pass@1}, while remaining best on \textit{pass@8}, indicating a minor trade-off between accuracy and exploration.

\subsection{Training Dynamics}

\Needspace{17\baselineskip}
\begin{wrapfigure}[15]{r}{0.55\linewidth}
\centering
\vspace{-20pt}
\includegraphics[width=\linewidth]{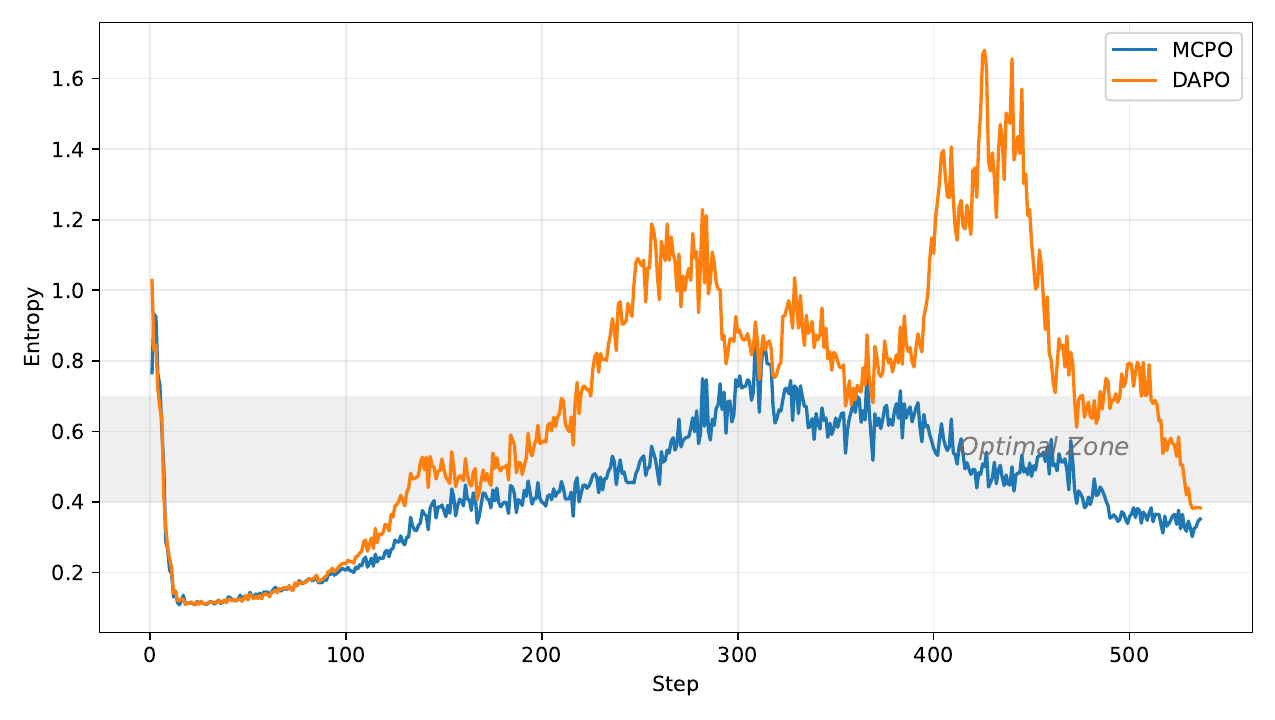}

\caption{Entropy dynamics comparison between MCPO and DAPO. The shaded region indicates the optimal entropy zone for exploration and exploitation balance ~\cite{wu2025quantile}.}
\label{fig:entropy}

\end{wrapfigure}
In this subsection, we introduce the several training dynamics of MCPO and DAPO on the Qwen3-8B-base model.
\vspace{0.5em}
\newline
\textbf{Entropy Dynamics.} 
The entropy dynamics during training is one of the most closely scrutinized aspects of RLVR methods. As shown in Fig.~\ref{fig:entropy}, the entropy dynamics during MCPO training is relatively moderate. No  entropy collapse phenomenon common in vanilla GRPO is observed during MCPO training. MCPO also mitigates  the phenomenon of entropy explosion~\cite{wu2025quantile} exhibited by DAPO. Entropy collapse hinders exploration and leads to repetition generation, entropy explosion induces excessive divergence that makes optimization unstable~\cite{wu2025quantile}.
\Needspace{16\baselineskip}
\begin{wrapfigure}[14]{r}{0.55\textwidth}
  \centering
  \vspace{-50pt} \includegraphics[width=\linewidth]{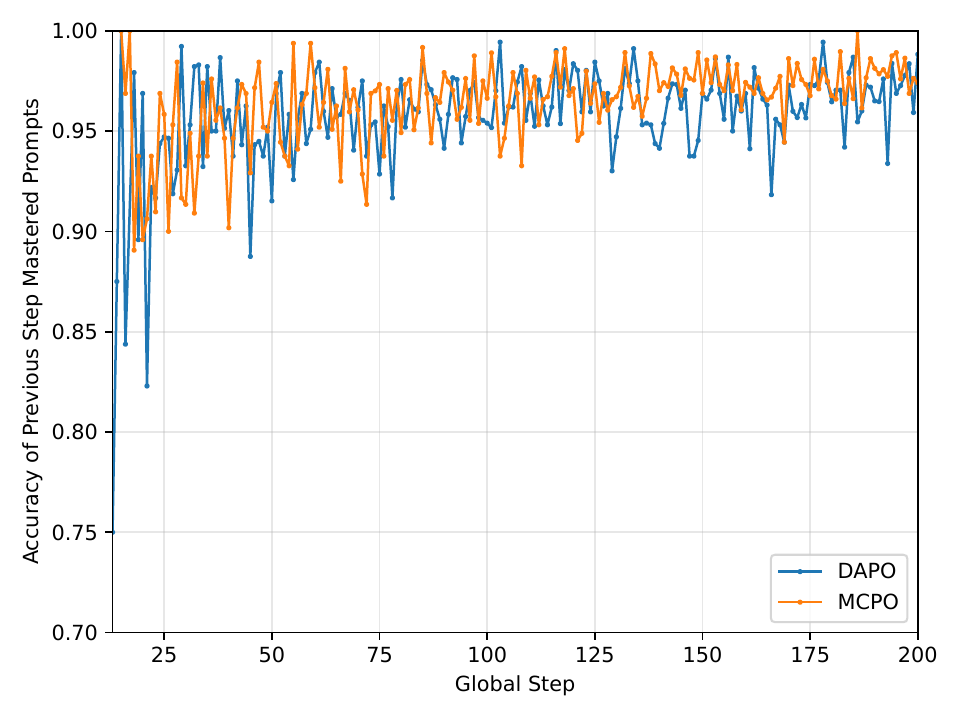}
  \vspace{-20pt}
  \caption{Comparison of mastered-prompt accuracy retention between DAPO and MCPO. MCPO generally maintains a higher accuracy floor with smaller fluctuations across training.}
  \label{fig:mcpo_master}
\end{wrapfigure}
\vspace{0.5em}
\noindent \textbf{More Mastered Prompts and Less All Wrong Prompts.} 
Beyond entropy, we further analyze statistics the fraction of mastered prompts and the fraction of all wrong prompts whose rollout groups are uniformly incorrect.  As shown in Fig.~\ref{fig:mastered-frac}, MCPO yields a higher mastered prompt fraction than DAPO throughout training, indicating that more prompts reach and stay in a fully correct regime. These trends align with the design of MCPO: the hinge-KL consolidation term explicitly bounds step-to-step drift on mastered prompts, preventing regression after mastery, while the query-weight balancing keeps majority-correct prompts influential so they can be pushed into full mastery instead of being progressively downweighted as rollout precision increases. Complementarily, Fig.~\ref{fig:all-wrong-frac} shows that MCPO consistently reduces the fraction of all wrong prompts, suggesting that fewer prompts remain stuck in complete failure. This counter-intuitive phenomenon is beyond our expectation, since MCPO is designed to primarily consolidate and preserve performance on mastered prompts and majority mastered prompts, rather than explicitly optimizing for hard prompts or all wrong prompts. 
This further corroborates the view that MCPO exhibits stronger exploratory capability from another perspective.
\vspace{0.5em}
\newline
\textbf{Accuracy Retention for Mastered Prompts.} During MCPO training, we track the mastered prompts in each step and compute their mean rollout accuracy under the next step policy. Except for replacing the training algorithm with MCPO, all other experimental procedures and settings are identical to those in Section~\ref{sec:observation}. As shown in the Fig.~\ref{fig:mcpo_master}, after introducing the hinge-KL loss term, the rollout accuracy on mastered prompts in the next step still decreases slightly. However, MCPO attains a higher accuracy floor than DAPO and exhibits smaller fluctuations. This provides evidence, to some extent, that MCPO leads to more stable knowledge retention during training. 
\section{Conclusion}
In this paper, we identify a mastery degradation phenomenon in Reinforcement Learning with Verifiable Rewards, where the model’s performance on previously mastered knowledge can regress during training. Such regression highlights the necessity of incorporating algorithms for explicit mastery consolidation. We propose two complementary consolidation strategies: (i) introducing a hinge-KL loss term on fully correct prompts to constrain policy drift between model policies between adjacent gradient step, and (ii) increasing the query weight of majority-correct prompts. Extensive experiments demonstrate that these consolidation mechanisms not only improve the accuracy performance of the model but also strengthen the exploration capability of the model. In future work, we will conduct a more in-depth theoretical investigation of mastery degradation and develop more efficient and robust mastery-consolidation methods.

\bibliographystyle{unsrt}
\bibliography{main}

\clearpage
\appendix
\numberwithin{equation}{section} 
\section{Appendix}
\subsection{Proof of the Query Weight in MCPO}
\label{app:mcpo_query_weight}

Assume binary rewards \(R(x,y)\in\{0,1\}\) and a non-degenerate rollout group for a prompt \(x\), i.e., \(0<p(x)<1\). Let the MCPO advantage be defined as Eq.~(12) in the main article, with the scaling function in Eq.~(13). After omitting the hinge-KL term, the reward-driven part of the MCPO objective can be written in the following discriminative form:

\begin{equation}
\begin{split}
\mathcal{J}_{\mathrm{MCPO}}(\theta)
=
\mathbb{E}_{x\sim\mathcal{D}}
\Big[
W(x)\cdot
\mathbb{E}_{y^{+}\sim \pi^{+}_{\theta_{\mathrm{old}}}(\cdot\mid x),\, y^{-}\sim \pi^{-}_{\theta_{\mathrm{old}}}(\cdot\mid x)}
\!\left[
s^{+}_{\theta}(y^{+},x)-s^{-}_{\theta}(y^{-},x)
\right]
\Big],
\end{split}
\label{eq:appendix_total_obj_with_Dhkl}
\end{equation}
where the query weight is
\begin{equation}
W(x)
=
\frac{\sqrt{p(x)(1-p(x))}}{\mathrm{scale}(p(x))}
=
\begin{cases}
\sqrt{p(x)(1-p(x))}, & 0<p(x)\le 0.5,\\[4pt]
0.5, & 0.5<p(x)<1.
\end{cases}
\label{eq:mcpo_query_weight}
\end{equation}

\noindent\textit{Proof.}
Write \(p=p(x)\) and \(s=\mathrm{scale}(p(x))\) for brevity. Since \(R(x,y)\in\{0,1\}\), the group mean and standard deviation are
\begin{equation}
\operatorname{mean}(\{R_i\}_{i=1}^G)=p,
\qquad
\operatorname{std}(\{R_i\}_{i=1}^G)=\sqrt{p(1-p)}.
\label{eq:mcpo_group_stats}
\end{equation}
Hence, for a correct response \(y^+\) and an incorrect response \(y^-\), the MCPO advantages are
\begin{align}
A^+_{\mathrm{MCPO}}(x)
&=
\frac{1-p}{\sqrt{p(1-p)}\,s}
=
\frac{1}{s}\sqrt{\frac{1-p}{p}},
\label{eq:mcpo_A_pos}\\
A^-_{\mathrm{MCPO}}(x)
&=
-\frac{p}{\sqrt{p(1-p)}\,s}
=
-\frac{1}{s}\sqrt{\frac{p}{1-p}}.
\label{eq:mcpo_A_neg}
\end{align}
Consider the token-normalized reward-driven surrogate
\begin{equation}
\mathcal{J}_{\mathrm{MCPO}}(\theta)
=
\mathbb{E}_{x\sim\mathcal{D}}
\mathbb{E}_{y\sim\pi_{\theta_{\mathrm{old}}}(\cdot\mid x)}
\left[
\frac{1}{|y|}\sum_{t=1}^{|y|}
f\!\left(r_t(x,y),A_{\mathrm{MCPO}}(x,y)\right)
\right].
\label{eq:mcpo_surrogate}
\end{equation}
Decomposing the expectation over responses into correct and incorrect groups yields:
\begin{equation}
\begin{aligned}
    \mathcal{J}_{\mathrm{MCPO}}(\theta) = \mathbb{E}_{x\sim\mathcal{D}} \Bigg[
    & p\, \mathbb{E}_{y^+\sim\pi^+_{\theta_{\mathrm{old}}}(\cdot\mid x)}
    \bigg[ \frac{1}{|y^+|}\sum_{t=1}^{|y^+|} f(r_t(x,y^+),A^+_{\mathrm{MCPO}}(x)) \bigg] \\
    & + (1-p)\, \mathbb{E}_{y^-\sim\pi^-_{\theta_{\mathrm{old}}}(\cdot\mid x)}
    \bigg[ \frac{1}{|y^-|}\sum_{t=1}^{|y^-|} f(r_t(x,y^-),A^-_{\mathrm{MCPO}}(x)) \bigg]
    \Bigg].
\end{aligned}
\label{eq:mcpo_surrogate_decomp}
\end{equation}
For the clipped GRPO-style surrogate, positive and negative advantages scale homogeneously. Specifically, for any \(c>0\),
\begin{equation}
f(r,c)=c\,f^+(r,1),
\qquad
f(r,-c)=-c\,f^-(r,1),
\label{eq:mcpo_homogeneity}
\end{equation}
where, for GRPO,
\begin{equation}
f^+(r,1)=\min(r,1+\epsilon),
\qquad
f^-(r,1)=\max(r,1-\epsilon).
\label{eq:mcpo_fpm}
\end{equation}
Accordingly, define
\begin{equation}
\begin{aligned}
s^+_{\theta}(y^+,x) &= \frac{1}{|y^+|}\sum_{t=1}^{|y^+|}f^+\!\left(r_t(x,y^+),1\right), \\
s^-_{\theta}(y^-,x) &= \frac{1}{|y^-|}\sum_{t=1}^{|y^-|}f^-\!\left(r_t(x,y^-),1\right).
\end{aligned}
\label{eq:mcpo_scores}
\end{equation}
Substituting Eqs.~\eqref{eq:mcpo_homogeneity} and \eqref{eq:mcpo_scores} into Eq.~\eqref{eq:mcpo_surrogate_decomp} gives
\begin{equation}
\begin{aligned}
\mathcal{J}_{\mathrm{MCPO}}(\theta) = \mathbb{E}_{x\sim\mathcal{D}} \bigg[ 
& pA^+_{\mathrm{MCPO}}(x)\mathbb{E}_{y^+\sim\pi^+_{\theta_{\mathrm{old}}}(\cdot\mid x)}[s^+_{\theta}(y^+,x)] \\
& - (1-p)|A^-_{\mathrm{MCPO}}(x)|\mathbb{E}_{y^-\sim\pi^-_{\theta_{\mathrm{old}}}(\cdot\mid x)}[s^-_{\theta}(y^-,x)] 
\bigg].
\end{aligned}
\label{eq:mcpo_coeff_form}
\end{equation}
The two coefficients coincide:
\begin{equation}
pA^+_{\mathrm{MCPO}}(x)
=
(1-p)|A^-_{\mathrm{MCPO}}(x)|
=
\frac{\sqrt{p(1-p)}}{s}.
\label{eq:mcpo_common_weight}
\end{equation}
Therefore,
\begin{equation}
\begin{aligned}
\mathcal{J}_{\mathrm{MCPO}}(\theta) = \mathbb{E}_{x\sim\mathcal{D}} \Bigg[ 
\frac{\sqrt{p(x)(1-p(x))}}{\mathrm{scale}(p(x))} \cdot \bigg( 
& \mathbb{E}_{y^+\sim\pi^+_{\theta_{\mathrm{old}}}(\cdot\mid x)}[s^+_{\theta}(y^+,x)] \\
& - \mathbb{E}_{y^-\sim\pi^-_{\theta_{\mathrm{old}}}(\cdot\mid x)}[s^-_{\theta}(y^-,x)] 
\bigg) 
\Bigg],
\end{aligned}
\end{equation}
which identifies
\begin{equation}
W(x)=\frac{\sqrt{p(x)(1-p(x))}}{\mathrm{scale}(p(x))}.
\end{equation}
Finally, substituting the definition of \(\mathrm{scale}(p(x))\) from Eq.~(13) in the main article yields
\begin{equation}
W(x)=
\begin{cases}
\sqrt{p(x)(1-p(x))}, & 0<p(x)\le 0.5,\\[4pt]
0.5, & 0.5<p(x)<1.
\end{cases}
\end{equation}
This proves Eq.~(14) in the main article. \hfill\(\square\)

\end{document}